\documentclass[sigconf]{acmart}
\AtBeginDocument{%
  }

\usepackage{multirow}


\setcopyright{acmlicensed}
\copyrightyear{2026}
\acmYear{2026}
\acmDOI{XXXXXXX.XXXXXXX}

\acmConference[]{MM' 26}{October 14-November 1, 2026}{Rio de Janeiro, Brazil}

\acmSubmissionID{4858}

\begin{document}

\title[GFR-SAM: Training-Free Referring Camouflaged Object Segmentation via Cross-Image Prompting]{GFR-SAM: Training-Free Referring Camouflaged Object Segmentation via Cross-Image Prompting}


\author{Yilong Yang}
\authornote{Equal contribution.}
\email{Yilong.Yang@xmu.edu.cn}
\affiliation{%
  \institution{Key Laboratory of Multimedia Trusted Perception and Efficient Computing, Ministry of Education of China, Xiamen University}
  \city{Xiamen}
  \country{China}
}

\author{Jianxin Tian}
\authornotemark[1]
\email{thetian666@stu.xmu.edu.cn}
\affiliation{%
  \institution{Key Laboratory of Multimedia Trusted Perception and Efficient Computing, Ministry of Education of China, Xiamen University}
  \city{Xiamen}
  \country{China}
}

\author{Shengchuan Zhang}
\email{zsc_2016@xmu.edu.cn}
\affiliation{%
  \institution{Key Laboratory of Multimedia Trusted Perception and Efficient Computing, Ministry of Education of China, Xiamen University}
  \city{Xiamen}
  \country{China}
}

\author{Liujuan Cao}
\authornote{Corresponding author.}
\email{caoliujuan@xmu.edu.cn}
\affiliation{%
  \institution{Key Laboratory of Multimedia Trusted Perception and Efficient Computing, Ministry of Education of China, Xiamen University}
  \city{Xiamen}
  \country{China}
}

\renewcommand{\shortauthors}{Yang et al.}

\begin{abstract}
Referring Camouflaged Object Detection (Ref-COD) requires segmenting hidden targets guided by reference cues. While supervised methods are annotation-heavy and training-free approaches via sparse point-prompting are sensitive to localization errors, we propose GFR-SAM, a robust three-stage training-free framework. GFR-SAM shifts the paradigm from fragile point-matching to a "Generate-Filter-Refine" pipeline. First, we introduce In-Context Exemplar-guided Segmentation, empowering SAM3 with cross-image inference to generate candidate masks via holistic visual exemplars, bypassing its native intra-image constraints. Second, a Region-Global Contrastive Filtering module ranks candidates through DINOv3-based prototypical alignment, effectively suppressing background distractors. Finally, a Geometric-Semantic Refinement module synergizes bounding box and text prompts to recover fine-grained boundaries and enhance instance recall. Evaluated on the R2C7K benchmark, GFR-SAM outperforms existing training-free methods by 8.7\% in weighted F-measure ($F_\beta^w$) and competes with supervised state-of-the-art counterparts. Ultimately, this work underscores the potential of unlocking SAM3's latent capability for cross-image In-Context prompting, establishing a robust, training-free paradigm that effectively bridges the gap between general-purpose foundation models and specialized, label-intensive perception tasks without the need for task-specific fine-tuning.
\end{abstract}




\maketitle

\section{Introduction}
Camouflaged object detection (COD) \cite{fan2021concealed} segments objects embedded in natural surroundings by mimicking background textures and patterns. Despite its progress, traditional COD lacks the flexibility to isolate specific targets in complex scenes. To bridge this gap, Referring Camouflaged Object Detection (Ref-COD) \cite{10848348R2CNET} has emerged, requiring models to segment targets guided by reference cues (e.g., text or image). This task is pivotal for diverse applications, including agricultural monitoring \cite{wang2024depth,rustia2020application}, search-and-rescue \cite{dousai2022detecting}, and defense surveillance \cite{liu2025multi,haider2025identification}.
\begin{figure}[!t]
    \centering
    \includegraphics[width=1.0\linewidth]{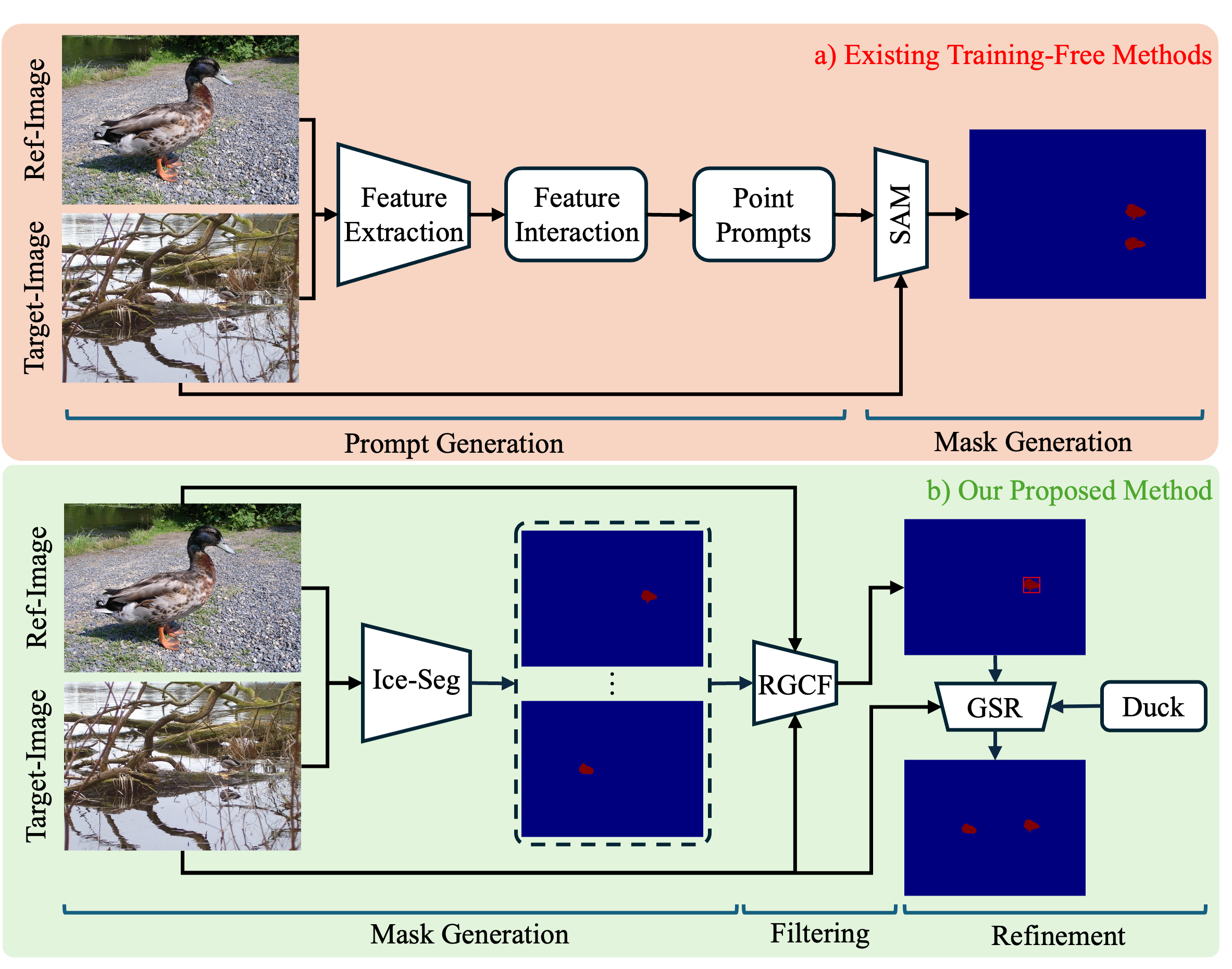}
    \caption{Comparison between the proposed GSR-SAM framework and prior training-free based segmentation methods.}
    \Description{Diagram comparing two image segmentation workflows. The top shows a less precise mask for a duck, resulting from existing methods. The bottom demonstrates our method, producing a highly accurate, refined duck mask through additional filtering and refinement stages.}
    \label{fig:fig1}
\end{figure}

Recent Ref-COD advancements predominantly rely on supervised learning \cite{10848348R2CNET, 11080234UAT, wu2025refonce, 11200287MLLM-RCOD}, which requires costly pixel-level annotations and suffers from overfitting to specific camouflage patterns, limiting generalization. To alleviate this dependency, training-free (zero-shot) methods like PerSAM \cite{zhang2023persam}, Matcher \cite{liu2023matcher}, IPSeg \cite{tang2025IPSeg}, and PPO \cite{liu2025plug-ppo} leverage foundation models (e.g., DINO \cite{caron2021emerging, oquab2023dinov2, simeoni2025dinov3} and SAM \cite{kirillov2023segany, ravi2024sam, carion2025sam}) via sparse point prompts. However, these point-based approaches exhibit a critical flaw in Ref-COD: extreme sensitivity to prompt location. Since foreground-background boundaries are indistinguishable, minor pixel-level shifts cause catastrophic segmentation failure. Furthermore, relying on local point-matching rather than holistic visual prototypes makes them highly susceptible to environmental distractors in complex scenes.

To overcome these limitations, we propose GFR-SAM, a novel three-stage training-free framework that shifts the paradigm from sensitive point-matching to a robust "Generate-Filter-Refine" pipeline. Unlike existing two-stage methods (point generation and mask generation) that suffer from prompt fragility, GFR-SAM decouples the process into three stages: In-Context Exemplar-guided Segmentation (Ice-Seg), Region-Global Contrastive Filtering (RGCF), and Geometric-Semantic Refinement (GSR). First, Ice-Seg leverages SAM3's Exemplar Encoder to extract a holistic visual prototype from the reference image, generating robust candidate masks and bypassing point-sensitivity issues. Second, RGCF ranks candidates via a semantic heatmap computed from cosine similarities between the global reference prototype and target patch embeddings, effectively suppressing background distractors. Finally, GSR combines the bounding box of the top-1 mask with text prompts, utilizing SAM3's native refinement capability to recover fine-grained boundaries. Consequently, GFR-SAM achieves state-of-the-art performance without parameter updates or pixel-level labels. The key contributions are summarized as follows:
\begin{itemize}
 \item We propose GFR-SAM, a training-free framework that decomposes Ref-COD into a "Generate-Filter-Refine" pipeline, balancing global target discovery with local boundary precision.
 \item We introduce In-Context Exemplar-guided Segmentation to empower SAM3 with cross-image inference, enabling robust guidance via external visual exemplars instead of fragile point prompts.
 \item We design a Region-Global Contrastive Filtering module leveraging DINOv3 prototypical alignment to rank candidate masks, effectively suppressing background distractors.
 \item We incorporate a Geometric-Semantic Refinement strategy that synergizes bounding box and text prompts, significantly enhancing recall in complex multi-object scenarios.
\end{itemize}

\section{Related Work}
\subsection{Referring Camouflaged Object Detection}
COD aims to segment objects that are inherently difficult to distinguish from their surroundings due to high visual resemblance, scale variations, 
and ambiguous appearance. While various strategies, such as multi-scale feature fusion \cite{huang2023FSPNet} and attention mechanisms \cite{zhang2022neighbor-connection} have been proposed to address these challenges, accurately identifying and segmenting a specific camouflaged object of user interest remains a significant hurdle. To address this critical limitation, the task of Ref-COD emerged. Ref-COD introduces external reference information, typically in the form of a textual description or a reference image, to guide the model in localizing and segmenting the target object.

The foundational work on Ref-COD, along with its inaugural dataset, was concurrently presented by \cite{10848348R2CNET}, which also proposed R2CNet as a baseline. R2CNet adopted a dual-branch architecture, designed to guide segmentation by learning a joint representation between the reference and target images. Subsequent research has explored diverse reference modalities, including text prompts , or salient exemplar object images as references \cite{11200287MLLM-RCOD}. From an architectural standpoint, UAT\cite{11080234UAT} pioneered the use of Transformer-based methods, effectively integrating reference semantics through cross-attention mechanisms and modeling uncertainty within the features. Furthermore, RFMNet\cite{wen2025RFMNet} advanced the field by enhancing local feature representation through multi-context feature fusion and overlapping window cross-attention techniques. More recently, RPMA\cite{liu2024RPMA} integrated reference information into general segmentation networks via a Cross Reference 
Adapter  and leveraged Reference-guided Dynamic Convolution  within the decoder to achieve flexible adaptation to varying references. Despite the impressive performance of these supervised methods, they inherently rely on extensive pixel-level annotations and expensive training phases, which limits their scalability and adaptability to diverse, real-world camouflage scenarios.

\subsection{Training free referring image segmentation}
Building upon the need for efficient adaptation, training-free segmentation has gained significant traction, primarily driven by the emergence of foundation models such as CLIP \cite{radford2021CLIP}, SAM \cite{kirillov2023segany} and DINO \cite{caron2021emerging}. CLIP aligns visual and textual representations via contrastive learning, while DINOv2 extracts strong image embeddings from unlabeled data. Leveraging these capabilities, a surge of training-free methods for Open-World Object Detection and Referring Image Segmentation (RIS) have emerged.
\begin{figure*}[t!] 
    \centering
    \includegraphics[width=\textwidth]{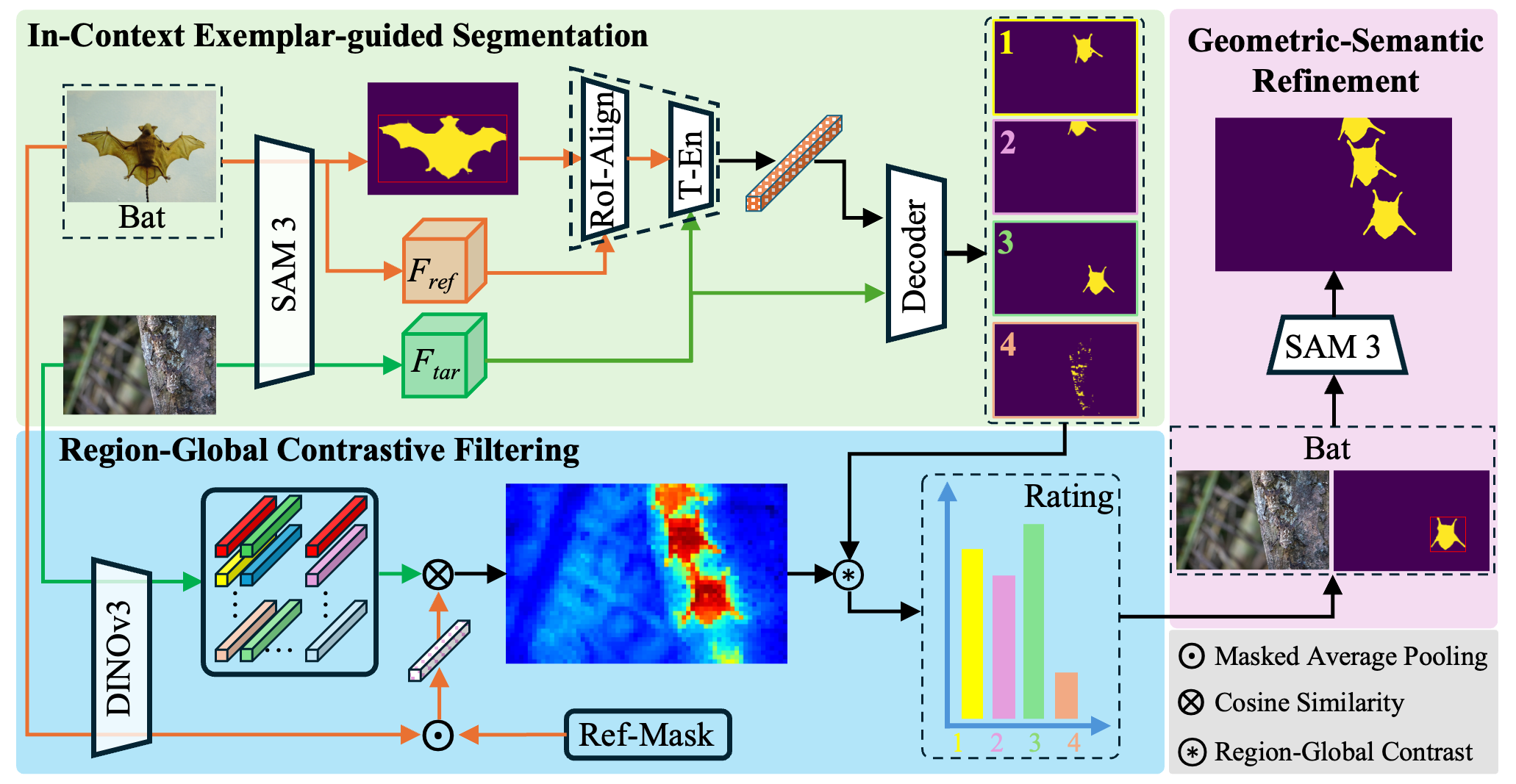} 
    \caption{The overall pipeline of the proposed Generate-Filtering-Refine Framework for Referring Object Segmentation. The framework consists of three stages: (a) The In-Context Exemplar-guided Segmentation (Ice-Seg) module generate candidate masks, prompted by the visual embedding of the reference image; (b) The Region-Global Contrastive Filtering (RGCF) module rates candidate mask and picks the most plausible one. (c) The Geometric-Semantic Refinement (GSR) module incorporates boundding box and text information to refine the initial segmentation. The colour coding of the arrows indicates the data flow of reference information (orange) and target information (green).}
    \label{fig:overview}
\end{figure*}

The dominant paradigm in this field leverages SAM by automatically generating point prompts based on reference images. Specifically, PerSAM \cite{zhang2023persam} utilizes vision foundation models to calculate location confidence maps from reference-target feature similarities, extracting positive and negative point prompts to explicitly guide SAM's attention. Similarly, Matcher \cite{liu2023matcher} performs bidirectional patch-level feature matching to sample a set of point prompts, followed by an instance-level mask filtering mechanism. To further enhance prompt accuracy, IPSeg \cite{tang2025IPSeg} fuses high-level semantic features from DINOv2 with low-level structural details from Stable Diffusion, identifying the most semantically consistent pixels to generate positive and negative point prompts. A critical limitation of the aforementioned point-based paradigm is that SAM is notoriously sensitive to the spatial distribution of point prompts; even a slight deviation, incorrect placement, or redundancy in the generated points can easily mislead the model, resulting in severe segmentation failures or incomplete masks. To overcome this inherent fragility, PPO (Point Prompt Optimizer) \cite{liu2025plug-ppo} introduces a plug-and-play optimization mechanism based on Deep Reinforcement Learning. Instead of relying solely on the initial heuristic matches, PPO constructs a dual-space heterogeneous graph and adaptively refines the distribution of the initial point prompts to ensure optimal segmentation.

Beyond point-based methods, VRP-SAM \cite{sun2024vrp} and VLP-SAM \cite{sakurai2025vision} utilize visual or multi-modal reference prompts for segmentation. However, their reliance on supervised training for feature alignment limits their utility in training-free scenarios. Recently, SAM3 \cite{carion2025sam} introduces Promptable Concept Segmentation, defining concept prompts as noun phrases, image exemplars, or their combinations. Despite this, its exemplar prompting is limited to intra-image scenario, where prompts must be defined within the target image. This constraint precludes in-context inference, rendering SAM3 unable to directly utilize independent reference images for Ref-COD. Thus, a specialized framework is required to bridge this gap for referring tasks with external visual cues.

\section{Method}
\subsection{Framework Overview}
The architecture of the proposed framework is illustrated in Figure \ref{fig:overview}, which consists of three synergistic stages designed to tackle the challenges of camouflaged object segmentation: In-Context Exemplar-guided Segmentation (Ice-Seg), Region-Global Contrastive Filtering (RGCF), and Geometric-Semantic Refinement (GSR).

Utilizing the Promptable Concept Segmentation of SAM3, Ice-Seg first extracts visual features for both the reference and target images. Then the patch embedding of the reference image, the reference mask, and the bounding box are injected into a Exemplar Encoder, where the deep interaction with the target features occurs to generate a robust reference prototype. Subsequently, the reference prototype is fed into the SAM3 decoder to generate a set of candidate masks for the target image. This stage effectively bypasses the native intra-image constraint of SAM3, enabling robust cross-image segmentation guided by holistic visual exemplars rather than fragile point prompts. However, due to high visual similarity between the camouflaged object and its background, the initial candidate mask set may include background clutter or semantically irrelevant objects. To robustly filter these candidates, we propose the Region-Global Contrastive Filtering (RGCF), which evaluates the semantic consistency of each candidate mask with the reference prototype, while also considering the global context of the target image. This is achieved through a contrastive scoring mechanism. Finally, the top-1 mask selected by RGCF is fed into the Geometric-Semantic Refinement (GSR) module. By synergizing bounding box prompts with text guidance, GSR leverages SAM3's native refinement capability to recover fine-grained boundaries, which proves particularly effective in enhancing segmentation recall and precision in complex multi-object scenarios.

\subsection{In-Context Exemplar-guided Segmentation}
To construct a robust reference prompt embedding $\mathbf{E}_{\text{prompt}}$, we adapt the Exemplar Encoder from the SAM3 foundation model. In this section, we first detail the standard encoding mechanism and then introduce our proposed In-Context Exemplar-guided Segmentation, which is specifically designed to refine reference object representation by mitigating background noise and spatial bias.
\subsubsection{Standard SAM3 Exemplar Encoder.} 
The original Exemplar Encoder in SAM3 is designed to process sparse geometric prompts (e.g., points and boxes) by fusing positional information with image-level features. In the standard implementation, a bounding box $b = [x_c, y_c, w, h]$ and its associated class label $l$ are encoded into a multi-modal prompt embedding $E_{prompt}$. This is typically achieved by aggregating four distinct components:
\begin{equation}
E_{prompt} = E_{box} + E_{pos} + E_{label} + E_{RoI}, E_{prompt}\in \mathbb{R}^d
\label{equ:e_prompt}
\end{equation}
where $d$ is the embedding dimension, $E_{box}$, $E_{pos}$, and $E_{label}$ denote the coordinate embedding, positional encoding, and label embedding, respectively. The $E_{RoI}$ denotes the regional-pooled visual features of the reference object obtained through RoI Align \cite{he2017mask}:
\begin{equation}
E_{RoI} = \mathcal{P}_{pool}(RoIAlign(F, b)),\quad E_{RoI} \in \mathbb{R}^d,
\label{eq:roi_pooling}
\end{equation}
where $F\in\mathbb{R}^{d\times N}$ is the flattened image features extracted by the SAM3 image encoder, $N$ is the number of patches. These components serve to provide geometric priors and semantic categories for the prompt. Subsecquently, $E_{prompt}$ serves as the query for a Transformer, interacting with the $F$ via attention mechanisms:
\begin{equation}
    \hat{E}_{prompt} = TransformerLayer(E_{prompt}, F).
    \label{eq:transformer}
\end{equation}
Finally, the updated prompt $\hat{E}_{prompt}$ is fed into the decoder to generate segmentation masks:
\begin{equation}
    \mathcal{M}= Decoder(F, \hat{E}_{prompt}),
    \label{eq:sam_decoder}
\end{equation}
where $\mathcal{M} = \{m_i\}_{i=1}^N$ is the set of candidate masks.
\subsubsection{In-Context Exemplar Encoder.} 
While standard SAM3 Exemplar Encoder is effective for intra-image instance recognition, it is not directly applicable to cross-image scenarios like Ref-COD. To address this, we modify the standard encoder to introduce an In-Context Exemplar Encoder.

Given the target image $I_{tar}$ and reference image $I_{ref}$, we first extract the visual features $F_{ref}$ and $F_{tar}$ using the SAM3 image encoder. Simultaneously, we obtain the corresponding binary mask $M_{ref}$ via SAM3's promptable concept segmentation. While standard RoI pooling (equation (\ref{equ:e_prompt})) effectively captures visual content within a defined region, it is suboptimal for reference object encoding in cross-image tasks. i.e., A bounding box $b_{ref}$ inevitably encompasses background pixels alongside the foreground object. Direct RoI pooling leading to contamination of the reference embedding $E_{prompt}$ with irrelevant features. To suppress background noise and ensure the embedding represents purely foreground semantics, we perform Mask-guided Spatial Gating (MSG) prior to pooling. By applying an element-wise Hadamard product $\odot$ between the feature map $F_{ref}$ and the binary mask $M_{ref}$, we effectively filter out non-target activations. This modified encoding process is formulated as:
\begin{equation}
    E^\prime_{RoI} = \mathcal{P}_{pool}(RoIAlign(F_{ref} \odot M_{ref}, b_{ref})), \quad E^\prime_{RoI}\in \mathbb{R}^d,
     \label{eq:modified_enc}
\end{equation}
where the spatial masking ensures that only the features within the precise boundary of the reference object contribute to the final prompt embedding. Furthermore, simply transferring the standard prompt embedding to a target image is flawed because the object's spatial location $b_{ref}$ in the reference image is usually not correspond to its location in the target image. Therefore, we discarde the coordinate-dependent components ($E_{box}$ and $E_{pos}$) and only retain the visual prototype $E^\prime_{RoI}$ as the prompt embedding for cross-image inference. To transfer this reference information to the target context, we modify equation (\ref{eq:transformer}) as follows:
\begin{equation}
    \hat{E}^{\prime}_{prompt} = TransformerLayer(
    E^{\prime}_{RoI},F_{tar}
    ).
     \label{eq:modified_transformer}
\end{equation}
By explicitly zeroing out background features (equation (\ref{eq:modified_enc})) within the RoI, removing position-specific geometric embeddings, and interacting with the target image features through a Transformer (equation (\ref{eq:modified_transformer})), this modified In-Context Exemplar Encoder forces the prompt embedding to rely exclusively on the object's visual signature. This significantly enhances the robustness of cross-image matching in the Ice-Seg framework. Finally, following equation (\ref{eq:sam_decoder}), the prompt embedding $\hat{E}^{\prime}_{prompt}$ and target image features $F_{tar}$ are fed into the SAM3 decoder for segmentation.
\subsection{Region-Global Contrastive Filtering}
Building upon the previous stage, the Ice-Seg module generates a set of primitive candidate masks $\mathcal{M} = \{m_i\}_{i=1}^N$ for the target image. However, this initial set is often noisy and contains false positives, especially in camouflaged scenarios where the target object blends into the background. To address this, the Region-Global Contrastive Filtering (RGCF) module is designed to robustly evaluate and filter these candidates, ensuring that only the most semantically consistent mask is selected for refinement. This is achieved through a contrastive scoring mechanism. The process involves computing a similarity heatmap between the reference prototype and the target image features, followed by a region-global contrast evaluation for each candidate mask.
\begin{figure*}[htpb] 
    \centering
    \includegraphics[width=\textwidth]{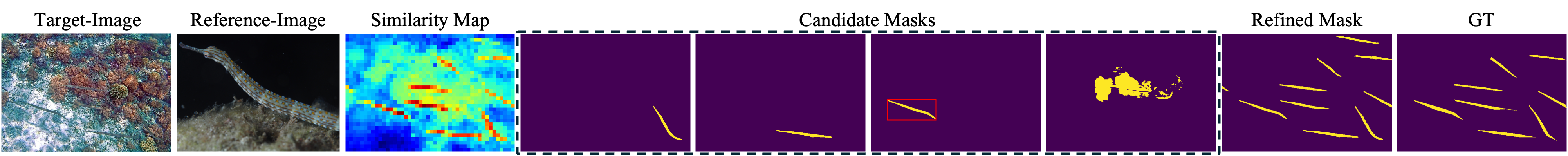} 
    \caption{Visualisation of intermediate and final segmentation of our method. The red box in the candidate mask indicates the top-1 mask selected by the RGCF module, which serves as the geometric prompt for the GSR that generates the final result.}
    \label{fig:intermediate}
\end{figure*}
\subsubsection{Similarity Heatmap Generation.} 
To bridge the semantic gap between the reference exemplar and the target image, we leverage the robust representation capabilities of the DINOv3 \cite{simeoni2025dinov3} backbone to generate a discriminative similarity heatmap. Specifically, given the reference image $I_{ref}$ and the target image $I_{tar}$, we first extract their patch-level feature maps using the shared DINOv3 encoder, denoted as $F_{ref} \in \mathbb{R}^{C \times H \times W}$ and $F_{tar} \in \mathbb{R}^{C \times H \times W}$, respectively (for notational simplicity, we assume the same size of the reference and target feature maps). To obtain a representative embedding of the reference object, we apply Mask Average Pooling (MAP) to the reference features $F_{ref}$ guided by the provided binary mask $M_{ref}$:
\begin{equation}
    P_{ref} = \frac{\sum_{x,y} (F_{ref}(x,y) \odot M_{ref}(x,y))}{\sum_{x,y} M_{ref}(x,y)},
\end{equation}
where $\odot$ denotes the Hadamard product, and $P_{ref} \in \mathbb{R}^C$ is the derived prototype feature representing the core visual attributes of the exemplar. Subsequently, we compute the dense correspondence between the prototype $P_{ref}$ and each spatial location in the target feature map $F_{tar}$. The similarity at each patch position $(x, y)$ is defined by the cosine similarity:
\begin{equation}
\mathcal{G}(x, y) = \frac{P_{ref} \cdot F_{tar}(x, y)}{\parallel P_{ref}\parallel \cdot \parallel F_{tar}(x, y)\parallel} + \epsilon,
\end{equation}
where $\mathcal{G} \in \mathbb{R}^{H \times W}$ is the resulting similarity heatmap, and $\epsilon$ is a small constant for numerical stability. The values in $\mathcal{G}$ represent the confidence of the presence of the reference object at each target patch. This heatmap $\mathcal{G}$ serves as the foundational response map for the subsequent Region-Global Contrastive Filtering stage.
\subsubsection{Region-Global Contrastive Score.} To ensure that the selected mask not only exhibits high visual similarity to the reference object but also stands out significantly from the global background context, we introduce a region-global contrastive (RGC) scoring mechanism. Specifically, given the similarity heatmap $\mathcal{G}$, we first compute the area-weighted average similarity (awas) for each candidate mask $m_i$:
\begin{equation}
S_{awas}(m_i) = \frac{\sum_{(x,y) \in \Omega_i} G(x,y)}{|\Omega_i|},
\end{equation}
where $\Omega_i$ denotes the set of pixels covered by the binary mask $m_i$, and $|\Omega_i|$ represents the mask area. While $S_{awas}$ measures the absolute local response, it may fail in cluttered environments where the background also exhibits high similarity. To suppress such distractors, we define the Region-Global Contrast score by normalizing the local response against the global average similarity $\bar{\mathcal{G}} = \frac{1}{HW} \sum \mathcal{G}(x,y)$:
\begin{equation}
  S_{RGC}(m_i) = \frac{S_{awas}(m_i)}{\bar{\mathcal{G}}},
\end{equation}
This ratio quantifies the relative saliency of each candidate mask: a higher RGC score indicates that the corresponding region is more discriminative against the global background. Finally, we rank all candidate masks $\mathcal{M}$ based on their RGC scores and select the top-1 mask as the optimal proposal for the subsequent refinement stage. Formally, the selected mask $m^*$ is defined as:
\begin{equation}
    m^* = \arg\max_{m_i \in \mathcal{M}} S_{RGC}(m_i)    
\end{equation}
\subsection{Geometric-Semantic Refinement}
While the RGCF module effectively localizes the most prominent target instance $m^*$, real-world applications often require exhaustive segmentation of all occurrences belonging to the same category. To enhance recall in multi-instance scenarios and suppress category-irrelevant noise, we utilize a Geometric-Semantic Co-prompting strategy. This approach leverages the Promptable Concept Segmentation capability of SAM3 by concurrently injecting geometric anchors and semantic constraints. Specifically, we derive a bounding box $b^*$ from the top-1 candidate $m^*$ to serve as the geometric prompt, providing a spatial reference for the target's scale and location. Simultaneously, the category name of the reference object is embedded as a semantic prompt $T_{sem}$ (e.g., "Bat") to activate the model's semantic-aware segmentation priors. The refinement process, which integrates these dual-modal prompts to identify all task-relevant regions in $I_{tar}$, is formulated as:
\begin{equation}
    \{m_i\}_{i=1}^N = \text{SAM3}(I_{tar}, b^*, T_{sem}, \tau),
\end{equation}
where $N$ denotes the number of detected instances that satisfy both the spatial-semantic constraints and the internal confidence requirement defined by $\tau$. Intuitively, a higher $\tau$ should suppress background noise; however, our empirical analysis reveals that for camouflaged scenarios, a lower $\tau$ is indispensable for recovering the complete structure of the target. Finally, these individual instance masks are aggregated via a pixel-wise union operation to yield the segmentation result:
\begin{equation}
    \text{Mask}_{\text{refined}} = \bigcup_{i=1}^{N} m_i.
\end{equation}
By synergizing geometric localization with semantic identity, this refinement stage effectively handles instance missing and ensures that the final output is both spatially precise and semantically pure. In Figure \ref{fig:intermediate}, we visualize the intermediate candidate masks generated by Ice-Seg, the top-1 mask selected by RGCF, and the final refined segmentation result, demonstrating the effectiveness and necessity of each stage in our GFR-SAM framework. i.e., while the initial candidates may contain noise and miss certain instances, the RGCF effectively identifies the most reliable proposal, and the GSR stage successfully recovers additional relevant instances, leading to a comprehensive and accurate segmentation outcome.
\begin{figure*}[!t] 
    \centering
    \includegraphics[width=\textwidth]{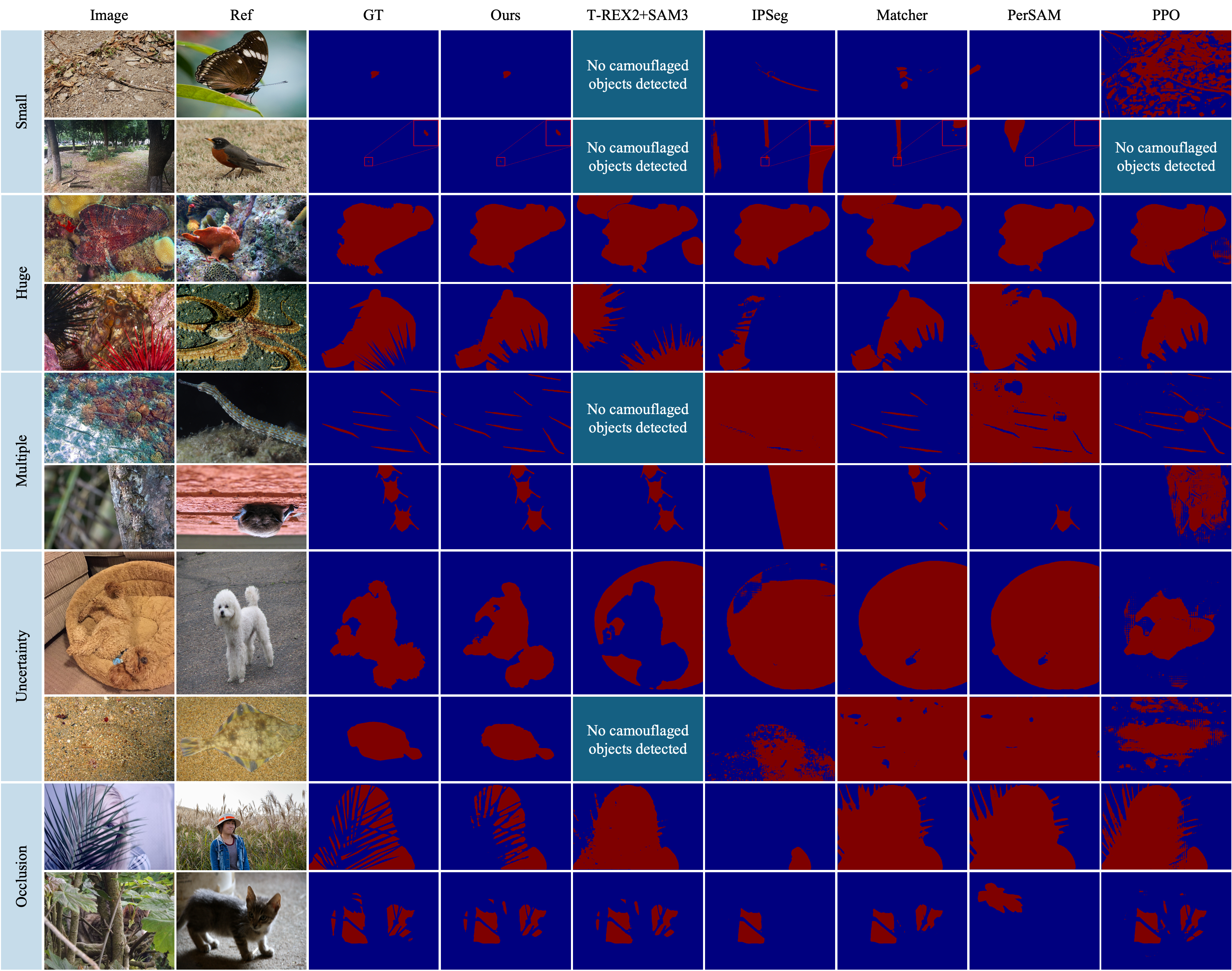} 
    \caption{Qualitative comparison of our GFR-SAM, T-REX2+SAM3~\cite{jiang2024trex2,carion2025sam},IPSeg~\cite{tang2025IPSeg}, Matcher~\cite{liu2023matcher}, PerSAM~\cite{zhang2023persam}, and PPO~\cite{liu2025plug-ppo}.}
    \label{fig:comparasion}
\end{figure*}
\begin{table*}[!t]
\centering
\caption{Comparison with recent state-of-the-art methods. The benchmark is performed on the R2C7K test set. The $N$ column denotes the number of reference image ("-" denotes that the model is prompted by the text prompt, i.e., the category name). "FS" and "TF" in the Sup column denote Fully Supervised and Training Free, respectively. The superscript $\dagger$ indicates that the method is point prompt-based. The best and second-best results are highlighted in bold and underlined, respectively.}
\label{tab:comparison}
\begin{tabular}{lccccccc|cccc|cccc}
\hline \hline
\multirow{2}{*}{Models} & \multirow{2}{*}{Venue} & \multirow{2}{*}{Sup} & \multirow{2}{*}{N}  & \multicolumn{4}{c}{Overall} & \multicolumn{4}{c}{Single-obj} & \multicolumn{4}{c}{Multi-obj} \\
\cmidrule(lr){5-8} \cmidrule(lr){9-12} \cmidrule(lr){13-16}
& & & & $S_m {\uparrow}$  & $\alpha$E ${\uparrow}$ & $F_\beta^w$ ${\uparrow}$ & M ${\downarrow}$  & $S_m{\uparrow}$ & $\alpha$E ${\uparrow}$ & $F_\beta^w$ ${\uparrow}$ & M ${\downarrow}$ & $S_m{\uparrow}$  & $\alpha$E ${\uparrow}$ & $F_\beta^w$ ${\uparrow}$ & M ${\downarrow}$  \\
\hline
R2CNET \cite{10848348R2CNET} & TPAMI 25 & FS & 5 & .805 & .879 & .669 & .036 & .810 & .880 & .674 & .035 & .747 & .872 & .602 & .046 \\
PFNet-S \cite{Mei_2021_CVPR_PFNet-S} & CVPR 21 & FS & 5 & .811 & .885 & .687 & .036 & .815 & .886 & .691 & .035 & .764 & .873 & .632 & .045 \\
UAT \cite{11080234UAT}  & TIP 25 & FS & 5 & .855 & .912 & .757 & .026 & .859 & .913 & .761 & .025 & .805 & .900 & .701 & .033 \\
RPMA-S \cite{liu2024RPMA} & ICME 24 & FS & 5 & .862 & .930 & .784 & .023 & .867 & .934 & .791 & .023 & .806 & .894 & .718 & .033 \\
RFMNet-S \cite{wen2025RFMNet} & ArXiv 25 & FS & 5 & .875 & .933 & .796 & .021 & .880 & .933 & .801 & .020 & .816 & .931 & .736 & .033 \\
MLLM-RCOD \cite{11200287MLLM-RCOD} & ICIVC 25 & FS & 5 & \underline{.890} & \textbf{.956} & \underline{.849} & \textbf{.017} & \underline{.894} & \textbf{.958} & \underline{.855} & \textbf{.017}  & .834 & \underline{.916} & .768 & .028 \\
RefOnce \cite{wu2025refonce} & ArXiv 25 & FS & 5 & \underline{.890} & .937 & .819 & \underline{.019} & \underline{.894} & .937 & .825 & \underline{.018} & \underline{.853} & \textbf{.930} & .767 & \textbf{.024} \\
\midrule
SAM3 \cite{carion2025sam3} & ICLR 26 & TF & -- & .799 & .783 & .664 & .034 & .805 & .789 & .673 & .033  & .749 & .728 & .583 & .045 \\
PerSAM$^\dagger$ \cite{zhang2023persam} & ICLR 24 & TF & 5 & .634 & .698 & .480 & .161 & .642 & .704 & .491 & .160 & .561 & .646 & .379 & .172 \\
Matcher$^\dagger $\cite{liu2023matcher} & ICLR 24 & TF & 5 & .839 & .901 & .778 & .040 & .842 & .902 & .781 & .040 & .810 & .896 & .752 & .043 \\
IPSeg$^\dagger$ \cite{tang2025IPSeg} & IJCV 25 & TF & 5 & .794 & .843 & .728 & .085 & .802 & .850 & .739 & .080 & .714 & .780 & .627 & .136 \\
PPO$^\dagger $\cite{liu2025plug-ppo} &  CVPR 25 & TF & 1 & .684 & .741 & .505 & .098 & .688 & .743 & .510 & .097  & .648 & .723 & .460 & .107  \\
T-REX2+SAM2 \cite{jiang2024trex2, ravi2024sam2} & ECCV 25 & TF& 5 & .802 & .791 & .672 & .034 & .806 & .794 & .678 & .033 & .767 & .763 & .621 & .040 \\
T-REX2+SAM3 \cite{jiang2024trex2, carion2025sam3} & ECCV 25 & TF& 5 & .804 & .793 & .675 & .033 & .806 & .794 & .678 & .033 & .789 & .788 & .649 & .031 \\
Ours & This work & TF & 3 & \textbf{.894} & \underline{.950} & \textbf{.865} & .021 & \textbf{.898} & \underline{.952} & \textbf{.869} & .020 & \textbf{.864} & \textbf{.930} & \textbf{.822} & \underline{.027} \\
\bottomrule
\end{tabular}
\end{table*}
\section{Experiments}
\subsection{Experiment Settings}
\paragraph{Datasets and Evaluations Metrics.} We evaluate our method on the R2C7K dataset \cite{10848348R2CNET}, which comprises 64 categories distributed across two subsets: the Cam subset (5,015 images containing camouflaged objects) and the Ref subset (1,600 salient reference images, 25 per category). Following R2CNet \cite{10848348R2CNET}, we report performance using four standard metrics: Structure Measure ($S_m$), Adaptive E-measure ($\alpha E$), Weighted F-measure ($F_\beta^w$), and Mean Absolute Error ($M$).

\subsection{Implementation details}
GFR-SAM is implemented in PyTorch using DINOv3 \cite{simeoni2025dinov3} and SAM3 \cite{carion2025sam}, with target category names as text prompts, $K=1$ in RGCF, and $\tau=0.35$ in GSR. All experiments run on a single NVIDIA RTX 3090 GPU (24GB). For a fair comparison, baselines utilize their official implementations and pre-trained weights. While most methods follow the R2C7K 5-shot protocol, PPO \cite{liu2025plug-ppo} and SAM3 are restricted to 1-shot and text-only prompting, respectively. Additionally, since the generic detector T-REX2 \cite{jiang2024trex2} outputs bounding boxes, its predictions are used as prompts for SAM2/SAM3 to generate masks.
\begin{table}[!t]
\centering
\caption{Ablation study on the use of Mask-guided Spatial Gating in the In-Context Exemplar Encoder. The RGCF and GSR modules are disabled in this experiment.}
\begin{tabular}{lcccc}
\hline \hline
Method & $S_m \uparrow$ & $\alpha E \uparrow$ & $F_\beta^w \uparrow$ & $M \downarrow$ \\
\hline
w/o MSG & .476 & .501 & .298 & .333 \\
with MSG & \textbf{.813} & \textbf{.850} & \textbf{.722} & \textbf{.064} \\
\hline \hline
\end{tabular}
\label{tab:ablation_ice_seg}
\end{table}
\subsection{Comparison with State-of-the-Art Methods}
\subsubsection{Qualitative Evaluation.} Figure \ref{fig:comparasion} visualizes the segmentation results of our method against leading competitors across diverse camouflaged scenarios. GFR-SAM consistently produces masks closely aligned with the ground truth, effectively capturing intricate boundaries under challenging conditions like extreme scales, low contrast, multi-instance, and occlusion. In contrast, baseline methods often suffer from over-segmentation, under-segmentation, or complete segmentation failure due to background ambiguity.
\subsubsection{Quantitative Evaluation.} Table \ref{tab:comparison} presents a comprehensive evaluation against recent state-of-the-art fully-supervised (FS) and training-free (TF) approaches on the R2C7K benchmark. Notably, GFR-SAM establishes a new state-of-the-art among TF paradigms, significantly outperforming robust foundation-model baselines (e.g., T-REX2+SAM3) and point-prompt methods (PerSAM, Matcher, IPSeg, PPO). Despite lacking parameter optimization, our method rivals or even surpasses recent FS specialists, achieving the highest structure measure ($S_m=0.894$) and weighted F-measure ($F_\beta^w=0.865$). We attribute this superiority to the In-Context Exemplar-guided Segmentation that bypasses point-prompt limitations via holistic visual exemplars. Furthermore, the RGCF module effectively suppresses background noise, while the GSR stage ensures comprehensive instance recall in complex scenes.
\subsection{Ablation Study}
In this section, we conduct extensive ablation study to validate the effectiveness of our framework's key components.
\begin{figure}[!t]
    \centering
    \includegraphics[width=1.0\linewidth]{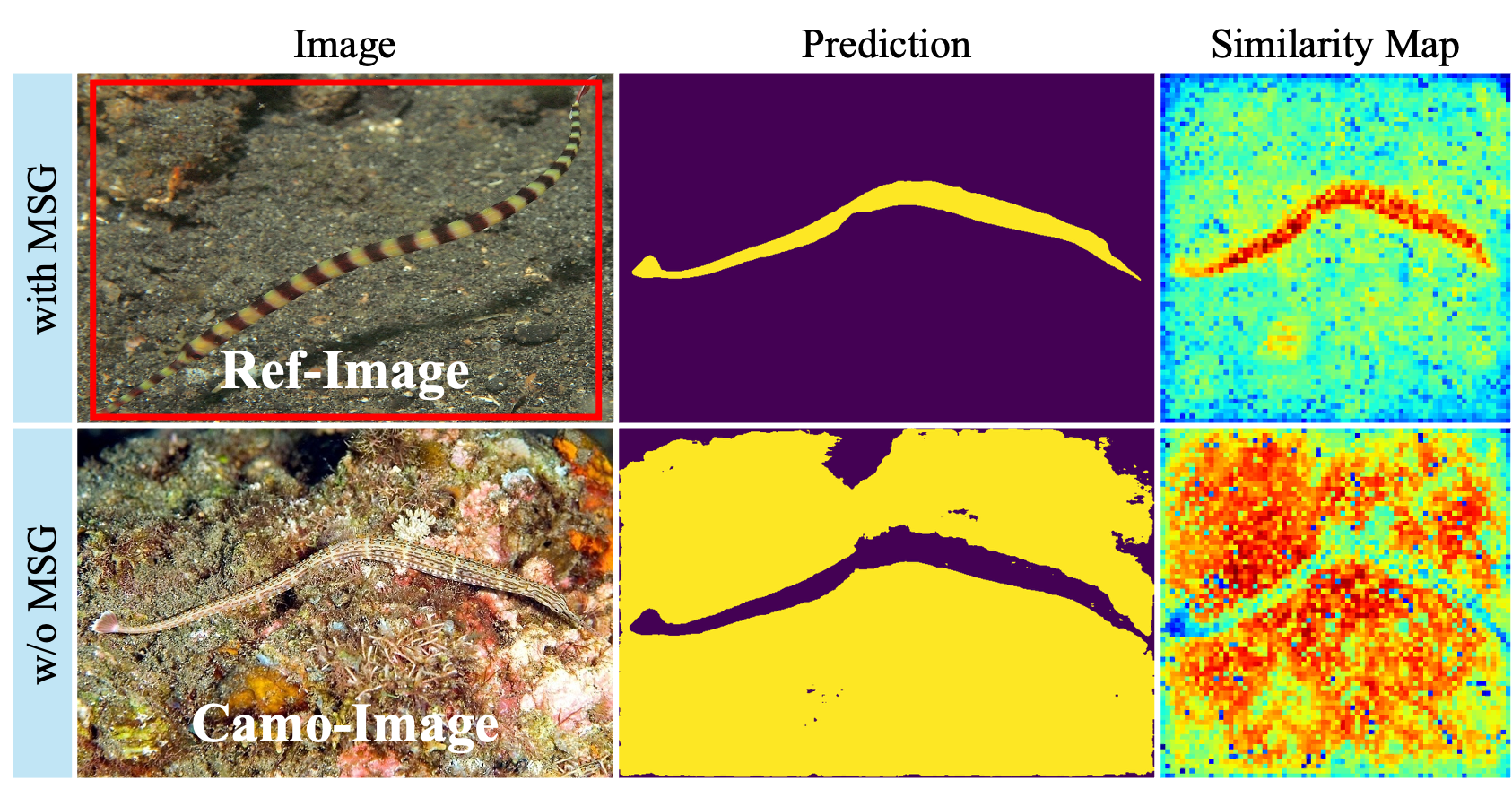}
    \caption{Visualization of similarity heatmaps between the target image features and reference prototype with and without Mask-guided Spatial Gating (MSG) in the Ice-Seg.}
    \label{fig:vis_enc}
\end{figure}
\subsubsection{Effectiveness of In-Context Exemplar Encoder.}
The core innovation of our In-Context Exemplar Encoder lies in the introduction of Mask-guided Spatial Gating (MSG) to suppress background features during the RoI pooling process. To validate the effectiveness of this design, we conduct an ablation study comparing the standard RoI pooling approach (equation (\ref{eq:roi_pooling})) against our modified version (equation (\ref{eq:modified_enc})). As shown in Table \ref{tab:ablation_ice_seg}, the inclusion of MSG leads to a dramatic performance boost across all metrics.  This substantial improvement underscores the critical role of Mask-guided Spatial Gating in enhancing the quality of the reference prototype by effectively filtering out irrelevant background noise, thereby enabling more accurate cross-image matching and segmentation. The visualization of the feature similarity in Figure \ref{fig:vis_enc} further illustrates this point, where the heatmap generated with MSG shows a more concentrated response around the target object, while the one without MSG exhibits widespread activations that include background regions, since the reference prototype derived from bounding box is contaminated by large amount of background features, thus activateing background regions instead of the target object. Notably, standalone Ice-Seg already outperforms point-based methods (e.g., PerSAM, IPseg, PPO), underscoring the effectiveness of our In-Context design in providing a robust initialization for Ref-COD.
\begin{figure}[!t]
    \centering
    \includegraphics[width=1.0\linewidth]{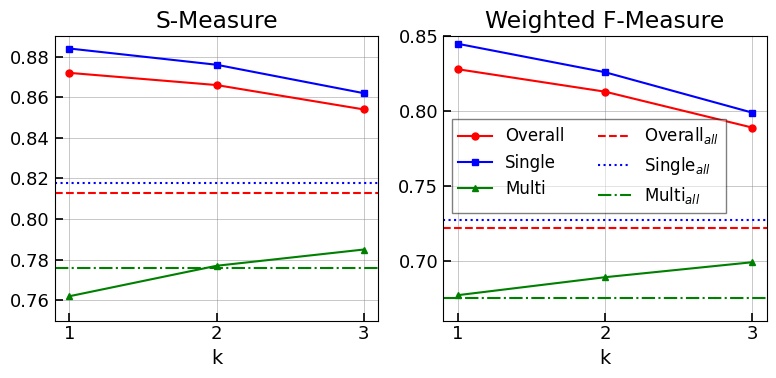}
    \caption{Ablation on RGCF with different values of $K$ for candidate mask selection and the baseline without selection.}
    \label{fig:fig6}
\end{figure}
\subsubsection{Effectiveness of Region-Global Contrative Filtering.}
We investigate the impact of the Top-$K$ constraint without the refinement module, as shown in Figure \ref{fig:fig6}. Retaining all mask candidates (all) inevitably introduces severe false positives from the background, resulting in a suboptimal overall score of 0.813 and 0.722 on $S_m$ and $F^w_\beta$ metrics, respectively. 
By enforcing a Top-$1$ selection, the performance improves significantly ($0.066\uparrow$ on $S_m$ and $0.118\uparrow$ on $F^w_\beta$) on the single-object subset, as it effectively filters out irrelevant candidates thus reducing false positives. However, increasing $K$ from 1 to 3 reveals an inherent dilemma prior to refinement: while a larger $K$ marginally benefits multi-object discovering ($S_m$ and $F^w_\beta$ metrics  increase by 0.23 and 0.22), it concurrently introduces noise that degrades single-object precision ($S_m$ and $F^w_\beta$ metrics drop by 0.22 and 0.46). This observation concludes that relying solely on Top-$K$ selection is insufficient for comprehensive robust segmentation, prompting the design of our subsequent refinement stage with $K=1$ as the explicit, high-confidence initialization.
\begin{table}[!t]
\centering
\caption{Ablation study on different visual feature encoders for the RGCF module. \textbf{Bold} indicates the best performance.}
\begin{tabular}{llcccc}
\hline \hline
Image Encoder & Backbone & $S_m \uparrow$ & $\alpha E \uparrow$ & $F_\beta^w \uparrow$ & $M \downarrow$ \\
\hline
\multirow{3}{*}{DINOv2} 
& ViT-S/14 & .890 & .945 & .856 & .021 \\
& ViT-B/14 & .891 & .945 & .858 & \textbf{.020} \\
& ViT-L/14 & .893 & .948 & .863 & \textbf{.020} \\
\hline
\multirow{3}{*}{DINOv3} 
& ViT-S/16 & .885 & .940 & .847 & .023 \\
& \textbf{ViT-B/16} & \textbf{.894} & \textbf{.950} & \textbf{.865} & .021 \\
& ViT-H+/16 & .889 & .942 & .854 & .023 \\
\hline
SAM3 & -- & .875 & .925 & .827 & .027 \\
\hline \hline
\end{tabular}
\label{tab:ablation_encoders}
\end{table}
\subsubsection{Ablation study on Feature Encoder.} Table \ref{tab:ablation_encoders} presents the ablation study on various visual feature encoders for the Region-Global Contrastive Filtering module on the R2C7K dataset. The comparisons demonstrate that DINOv3 with ViT-B/16 backbone consistently delivers superior performance across most metrics. It achieved the highest $S_m$, $\alpha E$, and $F_\beta^w$ while maintaining competitive $M$ values. The SAM3 image encoder also provides a strong baseline, but it falls short compared to the DINO series. We argue that the superior performance of DINOv3 can be attributed to its enhanced representation capabilities, which are crucial for capturing the subtle visual cues necessary for distinguishing camouflaged objects from their backgrounds. However, the use of SAM3 image encoder could be beneficial in scenarios where computational resources are limited, as it still offers reasonable performance without the need for additional feature extraction steps.
\subsubsection{Parameter Sensitivity Analysis on Confidence Threshold}
\begin{figure}[!t]
    \centering
    \includegraphics[width=1.0\linewidth]{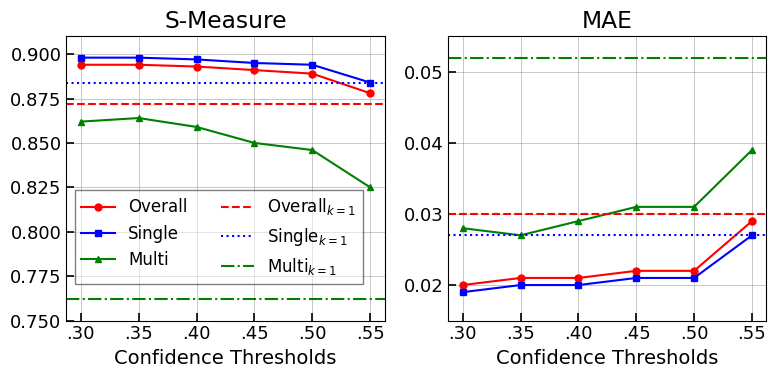}
    \caption{Ablation on varying confidence thresholds. The subscripts "k=1" denote that the measurements are based on the Top-1 candidate mask without refinement.}
    \label{fig:fig4}
\end{figure}

We conduct a sensitivity analysis on the confidence threshold ($\tau$) to investigate its impact on the GSR module. As shown in Figure \ref{fig:fig4}, performance variations are evaluated as $\tau$ sweeps from $0.30$ to $0.55$. A key observation is that smaller thresholds significantly outperform stricter ones in the Ref-COD task. Specifically, we find that a relaxed threshold (e.g., $\tau=0.35$) is indispensable for capturing camouflaged instances, which inherently yield weaker visual responses and lower confidence scores than generic objects.As $\tau$ increases beyond $0.5$, all performance metrics exhibit a sharp decline, indicating a "recall collapse" where SAM3 erroneously discards subtle true positive fragments. While a lower threshold may introduce marginal background noise, the resulting gains in structural integrity and multi-instance recall far outweigh the slight increase in MAE. By maintaining a lower $\tau$, GSR effectively "unlocks" SAM3's latent capacity to recover fragmented or low-contrast target regions that would otherwise be suppressed. Consequently, $\tau=0.35$ is selected as the optimal setting to ensure maximum target coverage in highly ambiguous environments.

Additional ablation on the number of shots and failure case analysis are provided in the supplementary material.

\section{Conclusion}
This paper presents GFR-SAM, a novel training-free Ref-COD framework that empowers SAM3 to transcend its native intra-image constraints. Our three-stage ``Generate-Filter-Refine'' pipeline seamlessly integrates In-Context Exemplar-guided Segmentation, Region-Global Contrastive Filtering, and Geometric-Semantic Refinement, resolving the prompt sensitivity and cross-image inference limits of existing methods. By mitigating background distractors and boosting instance recall, GFR-SAM achieves state-of-the-art training-free performance on the R2C7K benchmark, even rivaling supervised specialists. Overall, GFR-SAM demonstrates the feasibility of unlocking SAM3's latent cross-image in-context prompting, delivering a plug-and-play, annotation-free solution for challenging perception tasks without any parameter optimization.

\section*{Acknowledgments}
This work was supported by the National Science and Technology Major Project (No. 2025YFE0113500), the National Science Fund for Distinguished Young Scholars (No. 62525605), the National Natural Science Foundation of China (No. U25B2066, No. U22B2051, and No. 62272401), and the Xiamen Municipal Science and Technology Bureau, China (3502ZC-QXT2024009).
\bibliographystyle{ACM-Reference-Format}
\bibliography{sample-base}

\newpage

\appendix
\begin{table}[!b]
\centering
\caption{Comparison between different prompting strategies. The best results are highlighted in bold.}
\label{tab:comparison2}
\begin{tabular}{lcccc}
\hline \hline
Prompting Strategy & $S_m {\uparrow}$  & $\alpha$E ${\uparrow}$ & $F_\beta^w$ ${\uparrow}$ & M ${\downarrow}$  \\
\hline
Semantic only & .799 & .783 & .664 & .034 \\
Geometric only & .886 & .941 & .852 & .026 \\
Geometric+Semantic & \textbf{.894} & \textbf{.950} & \textbf{.865} & \textbf{.021} \\
\bottomrule
\end{tabular}
\end{table}
\section{Additional Ablation}
\subsection{Ablation on Prompting Strategies}\label{sec:gsr_ablation}
The Geometric-Semantic Co-Prompting strategy in our GSR module synergistically combines spatial cues from the geometric prompt (bounding box) with category-level semantics from the text prompt to guide SAM3's segmentation. To validate the contribution of co-prompting, we evaluate the performance of three prompting strategies: 1) Geometric-Only Prompting, which uses only the bounding box without any semantic input; 2) Semantic-Only Prompting, which relies solely on the category name without spatial guidance; and 3) the full Geometric-Semantic Co-Prompting approach. As shown in Table \ref{tab:comparison2}, the Geometric-Semantic Co-Prompting strategy significantly outperforms the semantic-only approach, demonstrating that spatial cues are crucial for localizing camouflaged objects. While the geometric-only prompting achieves a strong baseline, the addition of semantic information further enhances performance across all metrics. This improvement can be attributed to the fact that the semantic prompt activates SAM3's class-aware priors, enabling it to better differentiate between visually similar background regions and true target instances. The results underscore the importance of integrating both geometric and semantic information to fully leverage SAM3's capabilities for Ref-COD tasks.

\subsection{Ablation on the Number of Shots.} 
\begin{figure}[t]
    \centering
    \includegraphics[width=1.0\linewidth]{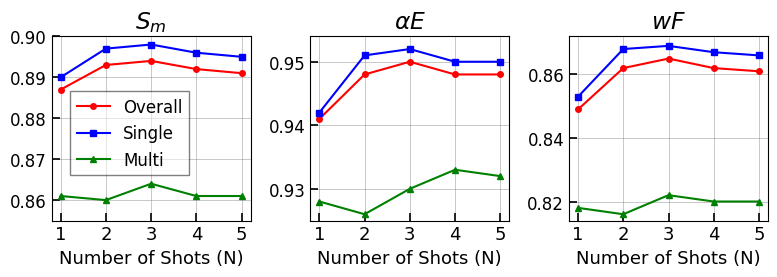}
    \caption{Ablation on the Number of Shots.}
    \label{fig:fig5}
\end{figure}

We investigate the influence of the support shot number ($N$) in Figure \ref{fig:fig5}. Unlike conventional feature-level averaging, our framework expands the candidate pool by independently generating proposals from each reference image. As $N$ increases from 1 to 3, all metrics consistently improve. This trend suggests that diverse visual perspectives enhance the probability of capturing the optimal target instance within a richer search space. However, performance saturates or slightly declines when $N \ge 4$. This saturation implies that an excessively large candidate pool introduces "distractor redundancy", where low-quality candidates from less relevant views may occasionally yield misleadingly high RGC scores. Consequently, we set $N=3$ as the default to strike an ideal balance between proposal diversity and selection precision.

\section{Failure Cases and Future Directions.}
\begin{figure}[t]
    \centering
    \includegraphics[width=1.0\linewidth]{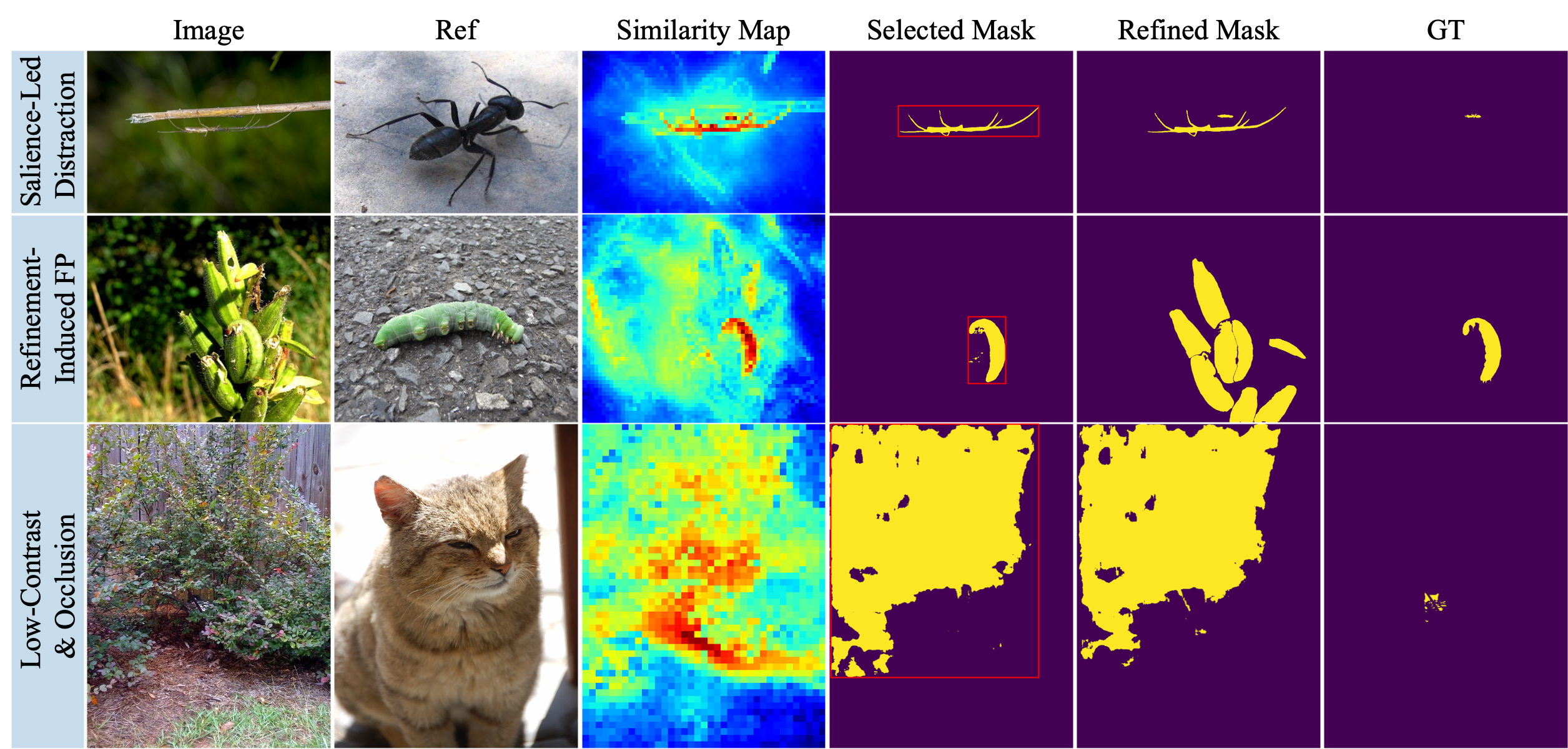}
    \caption{Qualitative visualization of failure cases. \label{fig:failure}}
\end{figure}

As illustrated in Figure. \ref{fig:failure}, our framework faces three primary challenges: 
\begin{itemize}
    \item \textbf{Salience-Led Distraction}: In scenarios where the scene contains non-camouflaged objects that are both visually similar to the target and highly salient, a "dominance" effect occurs. In these cases, the Ice-Seg stage fails to localize the camouflaged target and instead erroneously identifies the conspicuous distractor as the primary candidate.
    \item \textbf{Refinement-Induced False Positives}: We observe instances where Ice-Seg correctly segments the target and RGCF successfully selects it as the Top-1 proposal. However, the subsequent GSR module—designed to boost recall—introduces false positives by over-segmenting visually analogous background regions, suggesting that the semantic-geometric co-prompting can occasionally be overly inclusive.
    \item \textbf{Detection Failure in Low Contrast}: For diminutive targets under extreme occlusion and low contrast, the failure originates at the initial Ice-Seg stage. When the visual signal is too weak to establish reliable correspondences, the module fails to produce any valid candidate masks, highlighting a limitation in handling heavily fragmented or near-invisible targets.
\end{itemize}
To alleviate these issues, future research could be directed towards: 1) designing adaptive refinement mechanisms that can dynamically adjust the inclusion criteria based on the confidence of the initial segmentation to reduce false positives; and 2) enhancing the robustness of the initial exemplar encoding and matching process, potentially through multi-scale feature aggregation, to better handle low-contrast and heavily occluded targets.

\end{document}